%% file: main.tex
\documentclass[10pt,twocolumn,letterpaper]{article}
\usepackage[pagenumbers]{cvpr}

\input{preamble}

\definecolor{cvprblue}{rgb}{0.21,0.49,0.74}
\usepackage[pagebackref,breaklinks,colorlinks,allcolors=cvprblue]{hyperref}
\usepackage{algpseudocode}
\usepackage{algorithm}

\def\paperID{13733} %
\def\confName{CVPR}
\def\confYear{2025}

\title{
Diff2Flow: Training Flow Matching Models via Diffusion Model Alignment
}

\author{Johannes Schusterbauer* \qquad Ming Gui* \qquad Frank Fundel \qquad Bj\"orn Ommer \\
CompVis @ LMU Munich, MCML
}

\usepackage{tabularx}
\usepackage{multirow}
\usepackage{adjustbox}
\usepackage{makecell}
\usepackage{float}
\usepackage{environ}
\newcolumntype{Y}{>{\centering\arraybackslash}X}

\usepackage{subcaption}

\newcommand\circled[1]{\raisebox{.5pt}{\textcircled{\raisebox{-.9pt}{#1}}}}

\newcommand{\ven}[1]{\tiny\color{gray}(#1)}

\begin{document}
\maketitle
\def\thefootnote{*}\footnotetext{Equal Contribution}
\input{sec/0_abstract}

\section{Introduction}
\label{sec:intro}

\input{sec/1_intro}

\section{Related Work}
\label{sec:related_work}
\input{sec/2_related_work}

\section{Method}
\label{sec:method}
\input{sec/3_method}

\section{Experiments}
\label{sec:experiments}

\input{sec/4_experiments}

\section{Conclusion}
\label{sec:conclusion}
\input{sec/5_conclusion}

\section*{Acknowledgement}
This project has been supported by the bidt project KLIMA-MEMES, the German Federal Ministry for Economic Affairs and Climate Action within the project ``NXT GEN AI METHODS – Generative Methoden für Perzeption, Prädiktion und Planung'', the project ``GeniusRobot'' (01IS24083), funded by the Federal Ministry of Education and Research (BMBF), and the German Research Foundation (DFG) project 421703927.
The authors acknowledge the Gauss Center for Supercomputing for providing compute through the NIC on JUWELS at JSC and the HPC resources supplied by the Erlangen National High Performance Computing Center (NHR@FAU funded by DFG project 440719683) under the NHR project JA-22883.
{
    \small
    \bibliographystyle{ieeenat_fullname}
    \bibliography{main}
}

\newpage
\input{sec/X_suppl}

\end{document}

%% file: preamble.tex
\usepackage{pgffor}
\usepackage{float}
\usepackage[normalem]{ulem}

%% file: sec/0_abstract.tex
\begin{abstract}
Diffusion models have revolutionized generative tasks through high-fidelity outputs, yet flow matching (FM) offers faster inference and empirical performance gains. However, current foundation FM models are computationally prohibitive for finetuning, while diffusion models like Stable Diffusion benefit from efficient architectures and ecosystem support. 
This work addresses the critical challenge of efficiently transferring knowledge from pre-trained diffusion models to flow matching.
We propose Diff2Flow, a novel framework that systematically bridges diffusion and FM paradigms by rescaling timesteps, aligning interpolants, and deriving FM-compatible velocity fields from diffusion predictions.
This alignment enables direct and efficient FM finetuning of diffusion priors with no extra computation overhead. Our experiments demonstrate that Diff2Flow outperforms na\"ive FM and diffusion finetuning particularly under parameter-efficient constraints, while achieving superior or competitive performance across diverse downstream tasks compared to state-of-the-art methods. We will release our code at \url{https://github.com/CompVis/diff2flow}.
\end{abstract}

%% file: sec/1_intro.tex
Recently, diffusion models \cite{ho2020denoising,song2020score} have gained substantial popularity due to their exceptional generative capabilities, which have redefined the boundaries of image generation \cite{peebles2023scalable-dit, saharia2022photorealistic-imagen, podellsdxl, rombach2022high-sd-ldm}. 
Among these, foundation models such as Stable Diffusion \cite{rombach2022high-sd-ldm} stand out, not only for their high-fidelity outputs but also for their useful representations and adaptability to downstream tasks, including depth estimation \cite{ke2023marigold, gui2024depthfm}, surface normal prediction \cite{du2023generative-intrinsiclora}, segmentation \cite{xu2023open}, and semantic correspondences \cite{tang2023emergent-dift, fundel2025distilldift}.

Meanwhile, \textit{Flow Matching} (FM) \cite{albergo2023stochastic, lipman2022flow, liu2022rectifiedflow} has emerged as a promising alternative, empirically offering faster inference and improved performance \cite{esser2024scaling-sdv3, ma2024sit}.
While state-of-the-art foundational models based on flow matching, such as \textit{Flux} \cite{flux2024} or \textit{SDv3} \cite{esser2024scaling-sdv3}, show remarkable generative capabilities, their large size ($>8$B parameters) requires high-end hardware for both training and inference, making them particularly computationally expensive. This makes fine-tuning impractical, especially in resource-constrained environments and significantly limits their practical adoption.
In contrast, Stable Diffusion's \cite{rombach2022high-sd-ldm} efficient architecture and widespread ecosystem make it a pragmatic choice.
This raises an important question: Can the knowledge captured by existing foundational diffusion models be efficiently transferred to a flow matching model? How can we bridge the gap between diffusion and flow matching with minimal additional training, leveraging both the pre-trained prior and the advantageous properties of flow matching?

This work explores the relationship between the diffusion and flow matching paradigms. Although both methods can be generalized under a common framework \cite{albergo2023stochastic, ma2024sit, kingma2023understanding}, the actual implementations differ in several key aspects, including the definition of interpolants between the Gaussian noise prior and data samples, timestep scaling, and training objectives. These differences make it difficult to directly use a pre-trained diffusion model as a starting prior for flow matching training, as the two paradigms are not inherently well aligned.

To address this challenge, we propose a novel framework that effectively ``warps" diffusion into flow matching, enabling seamless knowledge transfer between the two paradigms. This requires re-scaling their respective timesteps, aligning their differing interpolant formulations, and deriving the velocity field required for flow matching from the diffusion model's predictions, based on its parameterization. 
By systematically establishing these correspondences, we enable a smooth transition between diffusion and flow matching. To this end, we introduce a training methodology termed \textit{Diff2Flow}, which initializes the flow matching model with a pretrained diffusion prior and directly finetunes it using the flow matching objective. Our analysis reveals that directly applying the FM loss to a diffusion model without incorporating our proposed adjustments significantly slows convergence and degrades overall model performance, whereas Diff2Flow provides a flexible and efficient approach with minimal finetuning overhead.

The benefits of this alignment become particularly evident when finetuning only a small subset of parameters. Under such constrained computational budgets, the performance gap between na\"ive FM finetuning and our alignment-aware approach becomes more pronounced. Specifically, we show that directly applying the FM objective with parameter efficient finetuning (PEFT) leads to very suboptimal performance.
Leveraging \textit{Diff2Flow} with its alignment-informed training strategy, PEFT enhances training efficiency and minimizes memory consumption while maintaining high performance.

We show the efficiency of our method on a diverse set of downstream tasks.
\circled{1} We show that when finetuning on a high-aesthetics dataset \cite{dai2023emu} on the text-to-image task, we outperform diffusion-based finetuning while converging significantly faster than na\"ively training with the flow matching objective.
This empirically demonstrates that smartly aligning timestep scaling and objective significantly facilitate learning, and additionally mitigate the zero-terminal SNR issue \cite{lin2024common} common in diffusion models, where full black or white images can't be generated.
\circled{2} When repurposing the model to generate images at resolutions different from the pre-trained sweet-spot resolution (similar to \cite{he2023scalecrafter}), our method achieves superior results compared to standard diffusion and flow matching fine-tuning.
\circled{3} We show that we can apply \textit{Reflow} \cite{liu2022rectifiedflow} using \textit{Diff2Flow} on a base diffusion model, a method that straightens sampling trajectories, resulting in faster inference speed. By rectifying the sampling trajectories of Stable Diffusion v1.5 we can generate images with as few as 2 sampling steps without consistency distillation \cite{song2023consistency}.
\circled{4} Finally, we demonstrate the effectiveness of our method on domain adaptation, leveraging the Stable Diffusion prior to predicting affine-invariant depth maps similar to Marigold \cite{ke2023marigold} and DepthFM \cite{gui2024depthfm}, achieving state-of-the-art results with reduced training time.

%% file: sec/2_related_work.tex
\paragraph{Diffusion and Flow Matching Models}
Diffusion models \cite{ho2020denoising, song2020score} have demonstrated wide-ranging capabilities in data synthesis, extending from image \cite{rombach2022high-sd-ldm, saharia2022photorealistic-imagen, ramesh2022hierarchical-dalle2, fuest2024diffusionrepresentation} and video \cite{ho2022video, blattmann2023align, guo2023animatediff} to audio generation \cite{liu2023audioldm} and beyond. While these models excel at producing high-fidelity outputs, they often require extensive sampling time, necessitating techniques like distillation \cite{sauer2025adversarial, meng2023distillation, salimans2022progressive, song2023consistency}, noise schedule optimization \cite{karras2022elucidating}, or training-free sampling \cite{song2021denoising-ddim, lu2022dpm} to achieve faster generation. Following diffusion models, flow matching models \cite{albergo2023stochastic, lipman2022flow, liu2022rectifiedflow} have gained attention due to their benefit of straighter probability paths. These models have been shown to perform competitively or surpass diffusion models in terms of both speed and quality \cite{ma2024sit, esser2024scaling-sdv3, schusterbauer2024fmboost}.

The relationship between diffusion and flow matching models has been explored in \citet{lee2024improving}, demonstrating that finetuning a flow matching model based on a pre-trained diffusion model is feasible. Our work extends this by explicitly formulating a method to map discrete diffusion trajectory to a continuous flow matching trajectory. Furthermore, we empirically demonstrate that our strategy works for various diffusion parameterization objectives and present results that go beyond reflow \cite{liu2022rectifiedflow}, achieving competitive performance on several benchmarks.

\paragraph{Parameter-efficient Finetuning}
As foundational models expand in size and complexity, parameter-efficient finetuning (PEFT) techniques such as Low-Rank Adaptation (LoRA) \cite{hu2022lora} offer viable alternatives to full model finetuning. Originally developed for large language models, LoRA and other PEFT techniques have since been applied effectively to diffusion models \cite{ryu2022low}, proving valuable across a broad range of tasks. 
These include domain-specific finetuning \cite{ye2023ip-adapter}, additional image conditioning \cite{stracke2024ctrloralter}, image editing \cite{gandikota2023concept}, distillation \cite{luo2023lcmlora} and among others. By updating only a low-rank decomposition of the weight matrix, LoRA reduces memory requirements and alleviate catastrophic forgetting \cite{biderman2024lora}, making it highly adaptable for targeted adjustments in large foundational models.

\paragraph{Knowledge in Generative Models}
Following the rise of text-to-image diffusion models \cite{rombach2022high-sd-ldm, nichol2022glide, podellsdxl, saharia2022photorealistic-imagen, ramesh2022hierarchical-dalle2}, numerous studies have explored methods to extract information from diffusion priors. Many works have effectively utilized the diffusion prior for tasks such as monocular depth estimation \cite{ke2023marigold,martingarcia2024diffusione2eft,fu2024geowizard, saxena2023monocular,gui2024depthfm, he2024lotus}, surface normal prediction \cite{du2023generative-intrinsiclora}, and semantic correspondences \cite{tang2023emergent-dift,luo2023diffusion-hyper, fundel2025distilldift}.
While these approaches predominantly focus on distilling information into another diffusion model, our work explores the transfer of this knowledge to flow matching models and hereby leverages the inherent advantages of flow matching, including more efficient inference and enhanced task performance.

%% file: sec/3_method.tex
\begin{figure}
    \centering
    \includegraphics[width=0.75\linewidth]{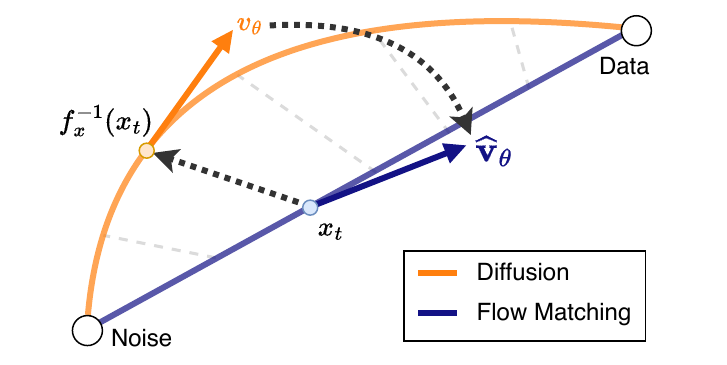}
    \caption{We introduce a novel finetuning technique to traverse between flow matching and diffusion, which enables effectively aligning the two processes with minimal additional training. 
    The interpolant on the flow matching trajectory is computed as a function $f$ of the diffusion timestep $t$, the sample $x$, and the associated diffusion coefficients. Our approach further enables a velocity prediction $\widehat{\mathbf{v}}$ regardless of the diffusion model's parameterization.
    }
    \label{fig:method:obj}
\end{figure}

In this section, we explain our algorithm and the process of aligning the trajectories. We begin by discussing the definition of diffusion and flow matching models. Following this, we describe how we traverse between the two trajectories as depicted in \cref{fig:method:obj} and use the diffusion model to generate a reasonable estimate of the flow velocity.

\subsection{Diffusion and Flow Matching}

\paragraph{Diffusion models} (DM) \cite{ho2020denoising, song2021denoising-ddim, song2020score} gradually diffuse data by typically adding noise to real data samples $x_0 \sim p_0(x_0)$ in a predefined discrete forward process, which can be characterized by
\begin{equation}
    p(x_t|x_0) = \mathcal{N}(x_t;\alpha_t x_0, \sigma_t^2 \mathbf{I}),
\end{equation}
where $\alpha_t$ and $\sigma_t$ are the predefined noise schedules, $t$ is the discrete timestep, and $x_t$ is the resulting noisy sample, often referred to as the interpolant. In the setting of variance preserving schedules $\sigma_t = \sqrt{1-\alpha_t^2}$, and the interpolant $x_t$ is then $x_t = \alpha_t x_0 + \sqrt{1-\alpha_t^2} \epsilon$ with $\epsilon \sim \mathcal{N} (0,\mathbf{I})$. A neural network, parameterized by $\theta$, is learned to reverse the forward process by gradually removing noise from $x_t$.
\citet{ho2020denoising} propose the simplified loss term
\begin{equation}
    \mathcal{L}_\text{simple} = \mathbb{E}_{t,\epsilon, x_0\sim p(x_0)} || \epsilon_t-\epsilon_\theta (x_t, t) ||^2,
    \label{eq:loss_simple}
\end{equation}
where predicting $\epsilon$ instead of $x_0$ is observed to yield better results and enhance convergence. There also exists the $v$-parameterization \cite{salimans2022progressive}, defined as
\begin{equation}
    v_t = \alpha_t \epsilon - \sigma_t x_0.
    \label{eq:v-param}
\end{equation}
\paragraph{Flow matching} (FM) models \cite{liu2022rectifiedflow, albergo2023stochastic, lipman2022flow}, another flexible class of generative models, utilizes the same idea of gradually deteriorating data samples and then synthesizing new data by reversing the process. We adopt the setting from \citet{lipman2022flow} where $x_0$ represents noise, and $x_1$ corresponds to data.
The interpolant on the continuous timesteps $t \in [0, 1]$ is defined as
\begin{equation}
    x_t = t x_1 + (1-t) x_0,
\end{equation}
where $x_0$ is typically a Gaussian noise sample.
The model $\mathbf{v}_\theta$ is trained to regress a vector field along the trajectory, following the linear path that points from $x_0$ to $x_1$ using the following objective:
\begin{equation}
    \mathcal{L}_\text{FM}=\mathbb{E}_{t,x_0\sim \mathcal{N}(0, \mathbf{I}),x_1\sim p(x_1)} || (x_1 - x_0) - \mathbf{v}_\theta(x_t, t) ||^2.
    \label{eq:fm_loss}
\end{equation}
Sampling from a flow matching model is accomplished by integrating over the trajectory of the learned ODE, e.g. using the forward Euler method with the update rule
\begin{equation}
    x_{t + t_\Delta} = x_{t} + t_\Delta \mathbf{v}_\theta (x_t, t)
    \qquad
    \forall t \in [0, 1),
    \label{eq:FM:sampling}
\end{equation}
with $t_\delta = 1/N$ and $N$ being the number of function evaluations (NFE). We generate a discrete sequence that approximates the path by advancing in small increments along the ODE. The trajectory's smoothness, or “straightness,” is critical, as it determines the number of steps required to achieve accurate results, trading off computational cost and fidelity. Straighter paths allow smaller $N$, whereas paths with higher curvature require more steps for simulating the ODE. 

\paragraph{Reflow}
In FM training, we randomly sample from a Gaussian distribution and form pairs with the data points to create the interpolant $x_t$. This random pairing leads to curved sampling trajectories, increasing the curvature of the ODE trajectories during inference \cite{tong2024improving}. To straighten these trajectories, Liu et al. \cite{liu2022rectifiedflow} propose \textit{Reflow}, which iteratively trains on samples obtained from a pre-trained flow matching model. Given an already trained ODE model $\mathbf{v}_\Phi$, Reflow first samples $x_0 \sim \mathcal{N}(0, \mathbf{I})$ and integrates along the ODE trajectory using \cref{eq:FM:sampling} to obtain a corresponding data sample $x_1$. This process generates image-noise pairings. Training on this paired data has been found to reduce inference cost by straightening the sampling trajectories \cite{liu2022rectifiedflow, liu2023instaflow, yan2024perflowpiecewiserectifiedflow}.

\begin{figure}
    \centering
    \begin{tabular}{cc}
        Normal timesteps $t_{{\text{DM}}}$ & Shifted timesteps $t_{\overline{\text{DM}}}$ \\
        \includegraphics[width=0.36\linewidth]{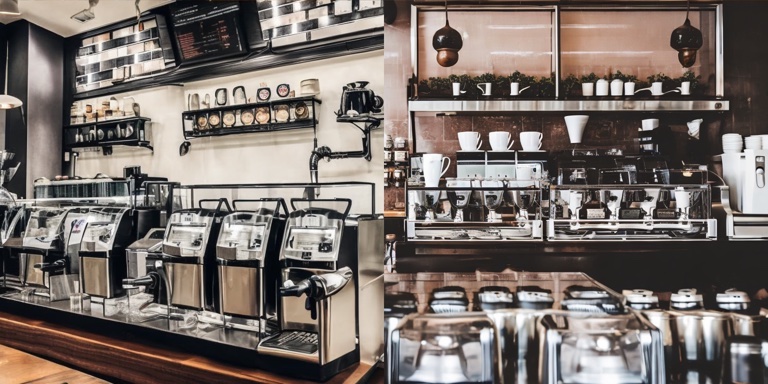} &
        \includegraphics[width=0.36\linewidth]{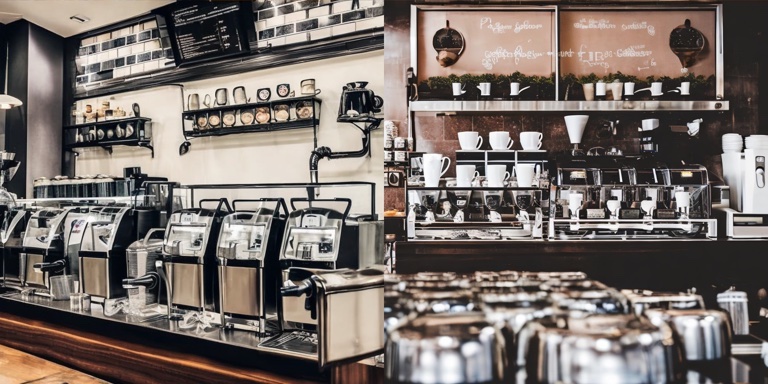}
    \end{tabular}
    \caption{Although non-integer-shifted DDIM timesteps are neither defined nor trained, generated images remain of high quality.}
    \label{fig:t2i:main:shifted_DDIM}
\end{figure}

\subsection{Diffusion Prior for Flow Matching Model}

Flow matching methods are known for straighter trajectories and more efficient inference \cite{gui2024depthfm, tong2024improving}. Hence, transforming the diffusion prior into a flow matching equivalent offers the potential for improvements in inference speed and performance \cite{ma2024sit, esser2024scaling-sdv3}. While previous research has successfully adapted diffusion priors for other downstream diffusion tasks, our approach combines these advantages with those of flow matching models.

Importantly, recent works on utilizing flow matching models to inherit a diffusion prior for downstream applications, such as DepthFM \cite{gui2024depthfm}, have shown significant promise in accelerating inference compared to diffusion models.
While this approach provides faster sampling than its diffusion model's counterpart, this inheritance method has limitations: DepthFM finetunes from a $v$-parameterized text-to-image diffusion model \cite{rombach2022high-sd-ldm} directly. This approach requires the flow matching model to ``warp" the $v$-parameterization to predict velocity, a different parameterization scheme. This introduces misalignments in the training objective, the interpolant, and the timesteps.

To address these limitations, we propose a novel strategy that enables a seamless transition between diffusion and flow matching trajectories. Our method minimizes misalignments and yields more accurate velocity predictions. In addition to monocular depth estimation, we demonstrate that our approach generalizes to other downstream tasks and offers a versatile and efficient solution for leveraging prior information from diffusion model.

\subsubsection{Traversing Between Trajectories}
Let us transform a foundation diffusion model's trajectory $x_t$ by applying two invertible operations, one $f_t$ that reparameterizes $t \in [0,T]$ and one $f_x$ that transforms $x$ to an alternate trajectory, defined by the following equation:
\begin{equation}
    \bar{x}(f_t) = f_x(x_{f_t}).
    \label{eq:traversing_definition}
\end{equation}
The boundary conditions need to satisfy $\bar{x}(f_0) = x_0$ and $\bar{x}(f_T) = x_T$ to ensure that the starting and target samples are the same. This is different to \citet{shaul2023bespoke} since our approach rectifies the trajectories, which requires the exact alignment of the starting and the target distribution.

In order to construct $f_t$ and $f_x$ reasonably, we need to probe into the diffusion and flow matching model and investigate how the interpolants are related. 
Let $x^\text{DM}$ represent a data sample on the diffusion trajectory and $x^\text{FM}$ a data sample on the flow matching trajectory. Similarly, $t_\text{DM}$ is the diffusion timestep and $t_\text{FM}$ is the flow matching timestep. Following the convention of diffusion and flow matching literature \cite{ho2020denoising,albergo2023stochastic,lipman2022flow,liu2022rectifiedflow,song2020score}, diffusion operates on discrete timesteps $t_\text{DM} \in \mathbb{Z}_{\geq0} \cap [0,T]$, where $x^\text{DM}_{t_\text{DM}=0}$ represents the data samples and $x^\text{DM}_{t_\text{DM}=T}$ corresponds to the Gaussian noise. In contrast, flow matching uses continuous timesteps $t_\text{FM} \in [0,1]$ with $x^\text{FM}_{t_\text{FM}=1}$ representing the data samples and $x^\text{FM}_{t_\text{FM}=0}$ representing the Gaussian noise. To summarize, the boundary condition enforces $x^\text{DM}_{t_\text{DM}=0}=x^\text{FM}_{t_\text{FM}=1}$ and $x^\text{DM}_{t_\text{DM}=T}=x^\text{FM}_{t_\text{FM}=0}$, while we seek two invertible mappings $f_t: [0, T] \rightarrow [0,1]$ which maps $t_\text{DM}$ to $t_\text{FM}$, and $f_x: [0,1] \times \mathbb{R}^d \rightarrow \mathbb{R}^d$ which maps $x^\text{DM}$ to $x^\text{FM}$.
The diffusion interpolant is
\begin{equation}
    x^\text{DM}_{t_\text{DM}} = \alpha_{t_\text{DM}}x^\text{DM}_0 + \sigma_{t_\text{DM}} x^\text{DM}_{T},
    \label{eq:interpolant-dm}
\end{equation}
where $\alpha_{t_\text{DM}}^2 + \sigma_{t_\text{DM}}^2 = 1$ for the variance-preserving schedule and $\alpha_{t_\text{DM}}=1,\ \forall t_\text{DM}$ with $\sigma_{t_\text{DM}}$ monotonically increasing over time for the variance-exploding schedule~\cite{ho2020denoising,song2020score}.

\noindent The flow matching interpolant is defined as
\begin{equation}
    x^\text{FM}_{t_\text{FM}} = t_\text{FM} x^\text{FM}_1 + (1-t_\text{FM}) x^\text{FM}_0.
    \label{eq:interpolant-fm}
\end{equation}
There are multiple ways to define $f_x$ so that the diffusion trajectory can be transformed and warped into a linear equation, one being
\begin{equation}
\begin{aligned}
    f_x(x_{t_\text{DM}}^\text{DM}) &= \frac{\alpha_{t_\text{DM}}}{\alpha_{t_\text{DM}}+\sigma_{t_\text{DM}}} x^\text{DM}_0 + \frac{\sigma_{t_\text{DM}}}{\alpha_{t_\text{DM}}+\sigma_{t_\text{DM}}} x^\text{DM}_{T} \\
    &= \frac{1}{\alpha_{t_\text{DM}}+\sigma_{t_\text{DM}}} x^\text{DM}_{t_\text{DM}},
    \label{eq:f_x_forward}
\end{aligned}
\end{equation}
yielding a scaled version of $x^\text{DM}_{t_\text{DM}}$.

To design $f_t$ as an invertible mapping between the discrete space $t_\text{DM}$ and the continuous space $t_\text{FM}$, we start by establishing an invertible transformation function that enables continuous interpolation between discrete points in $t_\text{DM}$ while aligning with $f_x$.
To ensure continuity and invertibility, we define $f_t$ as a piecewise linear function and interpolate between these discrete points. Let $t_{\overline{\text{DM}}} \in [0, T]$ be a continuous interpolated time space of the discrete $t_\text{DM}$. 
The correlation between \cref{eq:interpolant-fm} and \cref{eq:f_x_forward} allows to calculate $t_\text{FM}$ via the diffusion coefficients:
\begin{equation}
    f_t(t_\text{DM}) = \frac{\alpha_{t_\text{DM}}}{\alpha_{t_\text{DM}}+\sigma_{t_\text{DM}}},
    \label{eq:f_t_forward}
\end{equation}
where in both variance-preserving and variance-exploding diffusion schedules $f_t(t_\text{DM})$ is monotonic.
While the diffusion coefficients $\alpha_t$ and $\sigma_t$ are not defined for all $t_{\overline{\text{DM}}}$, we interpolate between the corresponding nearest neighbors. Let $t_{\text{DM}_1}$ and $t_{\text{DM}_2}$ be the nearest neighbors of $t_{\overline{\text{DM}}}$, and we design $f_t$ to be just as a linear interpolation in-between.
Interestingly, we find that although the continuous values of $t_{\overline{\text{DM}}}$ were not trained during the foundation model's training, direct inference with these values still produces reasonable results, as shown in \cref{fig:t2i:main:shifted_DDIM}. Specifically, instead of performing DDIM inference at discrete timesteps $t_\text{DDIM} \in \mathbb{Z}_{\geq0} \cap [0, 1000]$, we do inference at $t_\text{DDIM} + 0.5$ with linearly interpolated $\alpha_t$ and $\sigma_t$. 
We hypothesize that timesteps represented as sinusoidal embeddings create a well-defined continuous time space. This observation lays the foundation for interpolating timestep embeddings and traversing trajectories.

With $t_{\overline{\text{DM}}}$ defined on the continuous domain, the inverse $f^{-1}_t(t_{{\text{FM}}})$ can be defined as follows. First, we find $t_{\text{DM}_1}$ and $t_{\text{DM}_2}$ in the discrete timesteps such that $f_t(t_{\text{DM}_1})$ and $f_t(t_{\text{DM}_2})$ are the two nearest neighbors of $t_\text{FM}$ with $f_t(t_{\text{DM}_1}) \leqslant t_\text{FM} \leqslant f_t(t_{\text{DM}_2})$. Next, we perform a linear interpolation between the two neighbors to reverse the mapping from $t_{{\text{FM}}}$ to $t_{\overline{\text{DM}}}$:
    \begin{equation}
        f^{-1}_t(t_{{\text{FM}}}) = t_{\text{DM}_1} + \frac{t_{\text{FM}}-f_t(t_{\text{DM}_1})}{f_t(t_{\text{DM}_2})-f_t(t_{\text{DM}_1})}(t_{\text{DM}_2}-t_{\text{DM}_1}).
        \label{eq:f_t_backward}
    \end{equation}

With $f_t(\cdot)$ bidirectionally defined as in \cref{eq:f_t_forward} and \cref{eq:f_t_backward}, we deduce the bidirectional mapping from the FM trajectory $x^\text{FM}$ back to the DM trajectory $x^\text{DM}$:
\begin{equation}
\begin{aligned}
    f_x^{-1}(x^\text{FM}_{t_\text{FM}})
    &= (\alpha_{f^{-1}_t(t_{{\text{FM}}})}+\sigma_{f^{-1}_t(t_{{\text{FM}}})}) x^\text{FM}_{t_\text{FM}}. \\
\end{aligned}
\label{eq:f_x_backward}
\end{equation}

\begin{algorithm}[b]
    \caption{Diff2Flow Training}\label{alg:training}
    \begin{algorithmic}[1]
        \Require Data Sample $x_1^\text{FM}$, Noise sample $x_0^\text{FM}$, Timestep $t_\text{FM} \in [0,1]$
        \State Compute the interpolant $x^\text{FM}_{t_\text{FM}}$ \Comment{\cref{eq:interpolant-fm}}
        \State Reparameterize $t_{{\text{FM}}}$ to $t_{\overline{\text{DM}}}$ using $f_t^{-1}$ \Comment{\cref{eq:f_t_backward}}
        \State Reparameterize $x^{{\text{FM}}}_{t_\text{FM}}$ to $x^{{\text{DM}}}_{t_\text{DM}}$ using $f_x^{-1}$ \Comment{\cref{eq:f_x_backward}}
        \State Approximate velocity prediction $\mathbf{v}_\theta$  \Comment{\cref{eq:objective_change}}
        \State Take gradient descent step on $\mathcal{L}_\text{FM}$ \Comment{\cref{eq:fm_loss}}
        \end{algorithmic}
\end{algorithm}

\begin{algorithm}[b]
    \caption{Diff2Flow Sampling}\label{alg:inference}
    \begin{algorithmic}[1]
        \Require Noise sample $x_0^\text{FM}$, Number of sampling steps $N$
        \For{t = $0, \frac{1}{N}, \cdots ,\frac{N-1}{N} $}
        \State Reparameterize $t_{{\text{FM}}}$ to $t_{\overline{\text{DM}}}$ using $f_t^{-1}$ \Comment{\cref{eq:f_t_backward}}
        \State Reparameterize $x^{{\text{FM}}}_{t_\text{FM}}$ to $x^{{\text{DM}}}_{t_\text{DM}}$ using $f_x^{-1}$ \Comment{\cref{eq:f_x_backward}}
        \State Approximate velocity prediction $\mathbf{v}_\theta$  \Comment{\cref{eq:objective_change}}
        \State $x_{t+1/N} \gets x_{t} + \frac{1}{N} \mathbf{v}_\theta$ \Comment{\cref{eq:FM:sampling}}
        \EndFor \\
        \Return $x_1$
    \end{algorithmic}
\end{algorithm}

\subsubsection{Unifying the Objectives}

While traversing between the diffusion and flow matching trajectories is explained in the previous chapter, it is still essential to unify the training and inference objectives. 
Prior works \cite{gui2024depthfm, yan2024perflowpiecewiserectifiedflow} fine-tuned the model to adapt the diffusion prior, originally predicting $\epsilon$ or $v$, to directly predict velocity. However, these approaches force the model to transition between different parameterizations, therefore requiring longer convergence times and also ultimately impacting model performance. In contrast, our method unites the objectives by leveraging the diffusion prior and utilizing the relationships between the parameterizations. We term this technique \textit{Objective change} as we incorporate both trajectory traversal and a principled objective realignment to facilitate velocity prediction based on the diffusion prior. This approach is broadly applicable across different parameterizations; here, we demonstrate its effectiveness using $v$-parameterization.

Let $v_\theta(x^\text{DM},{t_\text{DM}})$ be the $v$-prediction (\cref{eq:v-param}) from a pre-trained diffusion model. Using the notation of \cref{eq:interpolant-dm}
\begin{equation}
    v_\theta(x^\text{DM},{t_\text{DM}}) = \alpha_{t_\text{DM}} x^\text{DM}_T - \sigma_{t_\text{DM}} x^\text{DM}_0
\end{equation}
where $x_0$ and the corresponding $x_T$ can be estimated as
\begin{equation}
\begin{aligned}
        \widehat{x^\text{DM}_0} &= \alpha_{t_\text{DM}} x^\text{DM}_{t_\text{DM}} - \sigma_{t_\text{DM}} v_\theta(x^\text{DM}_{t_\text{DM}},{t_\text{DM}}) \\
        \widehat{x^\text{DM}_T} &= \alpha_{t_\text{DM}} v_\theta(x^\text{DM}_{t_\text{DM}},{t_\text{DM}}) - \sigma_{t_\text{DM}} x^\text{DM}_{t_\text{DM}}, \\
\end{aligned}
\end{equation}
with the wide hat symbol indicating an estimate predicted by the model.
An approximation for the velocity $\mathbf{v}$ at the corresponding FM data point $x^\text{FM}$ is then formulated with respect to the boundary conditions defined in \cref{eq:traversing_definition}:
\begin{equation}
     \begin{aligned}
         \mathbf{v}_\theta(x^\text{FM}, t_\text{FM}) &= \widehat{x^\text{FM}_1} - \widehat{x^\text{FM}_0} \\
         &= \widehat{x^\text{DM}_0} - \widehat{x^\text{DM}_T} \\
         &= ({\alpha_{t_\text{DM}} - \sigma{t_\text{DM}}}) (x^\text{DM}_{t_\text{DM}} - v_\theta(x^\text{DM}_{t_\text{DM}},{t_\text{DM}})).
     \end{aligned}
     \label{eq:objective_change}
 \end{equation}
While we demonstrate the transition specifically from $v$-parameterization to velocity, this approach is versatile and can be readily applied to other parameterizations, such as $\epsilon$-parameterization, by following the same method outlined above.
\cref{alg:training} and \cref{alg:inference} summarize this approach in more detail. In summary, our main objective is to estimate the velocity using the prediction on the diffusion trajectory. We use standard Euler steps for inference.

\subsection{Parameter-efficient Finetuning}

Prior works \cite{luo2023lcmlora, ryu2022low} have demonstrated that parameter-efficient finetuning (PEFT) can be advantageous in tasks like domain adaptation and model distillation for generative models. PEFT reduces the number of parameters requiring updates, lowers memory demands during training, and theoretically mitigates catastrophic forgetting \cite{hu2022lora, biderman2024lora}. In our approach, we utilize Low-Rank Adaptation (LoRA) \cite{hu2022lora}, which achieves parameter efficiency by freezing the main model weights and updating only a low-rank decomposition of the weight matrices. Given a weight matrix $W_0 \in \mathbb{R}^{d\times k}$, we freeze $W$ and contrain the matrix's update with a low-rank decomposition $W_0 +\Delta W = W_0 + BA$, where $B \in \mathbb{R}^{d\times r}$ and $A \in \mathbb{R}^{r\times k}$ with the rank $r \leq \min(d,k)$. The modified forward pass for an input $x$ is then $h = W_0x+\Delta W x = W_0x+BAx.$
We observe that LoRA does not work out of the box when training a diffusion model with a flow matching objective, likely due to the shift required to adapt to a new parameterization scheme. In other words, LoRA struggles with the significant adjustment required when reconfiguring the model from a diffusion-based parameterization to a velocity prediction paradigm. This observation also underscores why na\"ive application of a flow matching objective to a pre-trained diffusion model is not effective; it forces the model to unlearn the old parameterization and switch to an entirely new one.
In contrast, LoRA performs well when incorporating our proposed objective change that aligns the trajectories. Here, LoRA provides an efficient balance, achieving strong results with minimal parameter updates and showing further improvements as more parameters are fine-tuned.

%% file: sec/4_experiments.tex
\subsection{Text-to-Image Synthesis}
\label{subsec:t2i}
\input{sec/4.1_txt2img}

\subsection{Rectifying the Trajectories}
\label{subsec:distill}
\input{sec/4.2_distillation}

\subsection{Monocular Depth Estimation}
\label{subsec:i2d}
\input{sec/4.3_img2depth}

%% file: sec/4.1_txt2img.tex
We start with the text-to-image task, the original task on which the diffusion prior model was trained. Specifically, we use Stable Diffusion 2.1 \cite{rombach2022high-sd-ldm} as our diffusion prior, pre-trained at a resolution of $768\times 768$, and fine-tune it to generate images at a lower resolution of $512\times 512$. This resolution shift allows us to investigate the effectiveness of different fine-tuning techniques in bridging resolution gaps.

\begin{figure*}[t]
    \centering
    \begin{subfigure}[b]{0.45\textwidth}
        \centering
        \includegraphics[height=0.12\textheight]{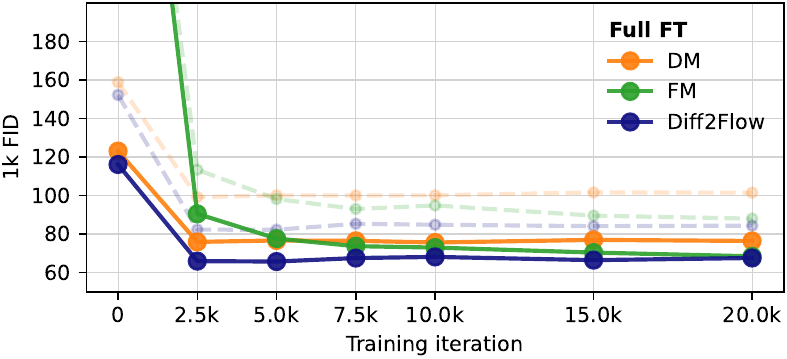}
        \caption{}
    \end{subfigure}
    \begin{subfigure}[b]{0.45\textwidth}
        \centering
        \includegraphics[height=0.12\textheight]{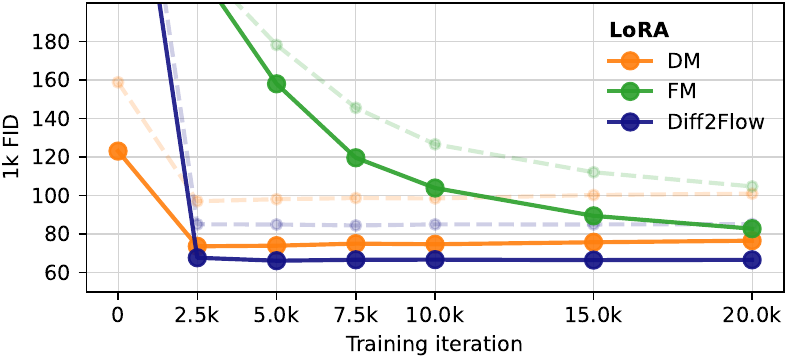}
        \caption{}
    \end{subfigure}
    \caption{Text-to-Image FID on the COCO 2017 \cite{lin2014microsoft_coco2017} validation dataset. Light curves indicate results without Classifier-free Guidance \cite{ho2021classifier_cfg}. \textbf{a)} We show that both FM and Diff2Flow converge to the same performance, given sufficient training and model capacity (full fine-tuning). However, our method removes the computational overhead from learning how to adjust the network output, resulting in faster convergence. \textbf{b)} Given limited modeling capacity (LoRA), the difference is more pronounced, where the FM model fails to close the gap to Diff2Flow, indicating that finetuning a portion of the parameters is enough to transfer a diffusion model to a flow matching model.}
    \label{fig:t2i:fid-training}
\end{figure*}

Starting from the SD2.1 prior, we continue training with different objectives on the LAION-Aesthetics dataset \cite{schuhmann2022laion} and evaluate model performance on COCO 2017 \cite{lin2014microsoft_coco2017}. \cref{fig:t2i:fid-training} shows that simply continuing diffusion training with an FM objective yields weak performance, especially when constraining model capacity with LoRA. In contrast, our method consistently surpasses continued diffusion training, both with and without classifier-free guidance, converging in as few as 2.5k iterations. Note that all models show poor FID results without training due to the resolution mismatch.
\input{figures/t2i_dmfmobj/img}
\input{figures/t2i_dmfmobj/img-zero-terminal-SNR}

We present qualitative results in \cref{fig:t2i:main:obj} and \cref{fig:t2i:main:zero-SNR}, where we use the same sampling hyperparameters and fix the initial Gaussian noise. A key issue that we also address in the transition from the diffusion model to flow matching is the problem of non-zero terminal SNR \cite{lin2024common}, which typically results in images generated by stable diffusion models having, on average, gray tones instead of true black or white. Our training paradigm solves this problem, illustrated in \cref{fig:t2i:main:zero-SNR}.

Furthermore, we investigate the scenario where diffusion training is resumed with our method without any changes in resolution or task objectives. Specifically, we initialize from an SD1.5 checkpoint and continue training on the LAION-Aesthetics dataset at a resolution of $512\times 512$. The corresponding evaluation metrics are reported in \cref{tab:additionalmetrics}.
Consistent with observations in SD3 \cite{esser2024scaling-sdv3}, the rectified trajectories of Diff2Flow yield superior performance compared to the baseline SD model. We show that converting a diffusion model to its flow-matching counterpart also leads to performance improvements within the same generative task.

\begin{table}
    \footnotesize
    \centering
    \setlength{\tabcolsep}{2.0pt}
    \begin{tabular}{@{}l@{\hskip .3in}c@{\hskip .1in}c@{\hskip .1in}c}
        \toprule
        Method & FID $\downarrow$ & CLIP $\uparrow$ & Aesthetics Score $\uparrow$  \\
        \midrule
        SD1.5                      & 56.77 & 26.34 & 5.32 \\
        SD1.5 (cont. training)     & 56.36 & 26.33 & 5.90 \\
        \textit{Diff2Flow}         & \textbf{52.80} & \textbf{26.54} & \textbf{5.99} \\
        \bottomrule
    \end{tabular}
    \caption{Eval on COCO-2017 5k with $25$ Euler/DDIM steps.}
    \label{tab:additionalmetrics}
\end{table}

%% file: figures/t2i_dmfmobj/img.tex
\begin{figure}
    \centering \small
    \setlength\tabcolsep{1pt}

    \newcommand{\quotes}[1]{``\textit{#1}''}

    \newcommand{\imagepng}[2]{
    \includegraphics[width=0.2\linewidth]{figures/t2i_dmfmobj/#1-#2.jpg}
    }

    \newcommand{\rowpng}[1]{
        \imagepng{DM}{#1} 
    &
        \imagepng{FM}{#1}
    &
        \imagepng{OBJ}{#1}
    }

    \begin{subfigure}{0.5\textwidth}
        \centering
        \begin{tabular}{lccc}
            & \footnotesize DM & \footnotesize FM & \footnotesize \textbf{Diff2Flow} \textit{(Ours)} \\
            \rotatebox{90}{\hspace{.5em} w/o LoRA} & \rowpng{1} \\[-1ex]
            & \multicolumn{3}{c}{\tiny \quotes{An astronaut walking on a green mountain path watching the sunset}} \\
            \rotatebox{90}{\hspace{.5em} w/ LoRA} & \rowpng{lora-2} \\[-1ex]
            & \multicolumn{3}{c}{\tiny \quotes{portrait of a cat covered in cloud of smoke, oil painting style}} \\[1ex]
        \end{tabular}
    \end{subfigure}
    \caption{Qualitative results for our finetuned Text-to-Image models using different objectives. \textbf{DM}, \textbf{FM} and \textbf{Diff2Flow} stand for diffusion finetuning, flow matching finetuning, and our method. Images are generated with the same seed, NFEs, and CFG $=4.0$.}
    \label{fig:t2i:main:obj}
\end{figure}

%% file: figures/t2i_dmfmobj/img-zero-terminal-SNR.tex
\begin{figure}
\center \small
    \setlength\tabcolsep{1pt}

    \newcommand{\quotes}[1]{``\textit{#1}''}

    \newcommand{\imagepng}[2]{
    \includegraphics[width=0.2\linewidth]{figures/t2i_dmfmobj/#1-#2.jpg}
    }

    \newcommand{\rowpng}[1]{
        \imagepng{DM}{#1} 
    &
        \imagepng{FM}{#1}
    &
        \imagepng{OBJ}{#1}
    }

    \begin{tabular}{lccc}
        &\footnotesize DM &\footnotesize FM & \footnotesize \textbf{Diff2Flow} \textit{(Ours)} \\
        \rotatebox{90}{\hspace{.5em} w/o LoRA} & \rowpng{3} \\
        & \multicolumn{3}{c}{\tiny \quotes{A fully white image with black borders}} 
        \\
        \rotatebox{90}{\hspace{.5em} w/ LoRA} & \rowpng{lora-1} \\ 
        & \multicolumn{3}{c}{\tiny \quotes{A fully white background with a gray circle in the center}} \\
    \end{tabular}
    \caption{Our finetuning objective, both with and without PEFT effectively addresses the issue of non-zero-terminal SNR of DMs. The generations align faithfully to the input prompts.}
    \label{fig:t2i:main:zero-SNR}
\end{figure}

%% file: sec/4.2_distillation.tex
\begin{table*}[t]
    \centering
    \input{tbl/i2d-sota}
    \caption{
        \textbf{Comparison to state-of-the-art depth estimation methods.} We compare discriminative (top) and generative models (bottom). Numbers obtained from \cite{martingarcia2024diffusione2eft}. $^{\dagger}$Metric3D v2~\cite{hu2024metric3dv2} was trained on ScanNet, so zero-shot evaluation is not possible. 
        We also show the results reproduced by \cite{martingarcia2024diffusione2eft} for GeoWizard. All of our results are generated with NFE$=10$ and an ensemble size of $4$.
    }
    \label{tab:i2d:sota}
\end{table*}

In addition to fine-tuning diffusion models for resolution changes, we can also optimize them for straighter sampling trajectories to enable faster generation speeds. A popular approach to this is \textit{Reflow} \cite{liu2022rectifiedflow}, which modifies the flow-matching training objective by replacing the random Gaussian noise terms in \cref{eq:fm_loss} with calculated noise derived from a pre-trained prior ODE to ensure that each noise-image pair is aligned according to the ODE. Quantitative results are presented in \cref{tab:reflow:sota}, where our method demonstrates competitive performance over various number of function evaluations (NFEs), including low NFEs such as 2 and 4, while only fine-tuning less than $7\%$ of the parameters. Qualitative results using reflow with 4 inference steps are shown in \cref{fig:reflow:cfg-ablation} and in the Appendix.

\begin{table}[t]
    \centering
    \input{tbl/reflow-sota}
    \caption{
    We fine-tune all our models with LoRA, and apply only 1-rectified flow according to \citet{liu2022rectifiedflow}. We get competitive results with state-of-the-art reflow models.
    }
    \label{tab:reflow:sota}
\end{table}

\begin{figure}[t]
    \centering
    \input{figures/txt2img/t2i_nfe-cfg}
    \caption{SD1.5 \cite{rombach2022high-sd-ldm} + Diff2Flow-Reflow, 4-step inference results.}
    \label{fig:reflow:cfg-ablation}
\end{figure}
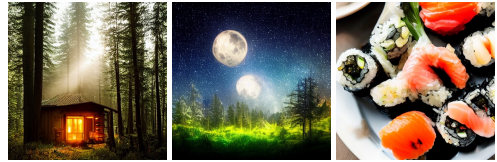

%% file: tbl/i2d-sota.tex
\footnotesize
\setlength{\tabcolsep}{2.0pt}
\begin{tabularx}{\textwidth}{
lcc
YYc
YYc
YYc
YYc
YY
}

\toprule

\multirow{2}{*}{Method} &
\multirow{2}{*}{\shortstack{\text{Training}\\\text{samples}}} &
&
\multicolumn{2}{c}{NYUv2 \cite{silberman12nyuv2}} & &
\multicolumn{2}{c}{KITTI \cite{geiger2013kitti}} & &
\multicolumn{2}{c}{ETH3D \cite{schops2017multiEth3d}} & &
\multicolumn{2}{c}{ScanNet \cite{dai2017scannet}} & &
\multicolumn{2}{c}{DIODE \cite{diode_dataset}} \\

\cmidrule{4-5}\cmidrule{7-8}\cmidrule{10-11}\cmidrule{13-14}\cmidrule{16-17}

&
&
&
AbsRel↓ & 
$\delta$1↑ & &
AbsRel↓ & 
$\delta$1↑ & &
AbsRel↓ & 
$\delta$1↑ & &
AbsRel↓ & 
$\delta$1↑ & &
AbsRel↓ & 
$\delta$1↑
\\

\midrule

Depth Anything \cite{depth_anything_v1}~\ven{CVPR '24} &
62M &&
 \textbf{4.3} & 
\textbf{98.1} & &
 7.6 & 
94.7 & &
12.7 & 
88.2 & &
--- &
--- & &
\underline{6.6} &
\underline{95.2}
\\

Depth Anything v2 \cite{depth_anything_v2}~\ven{arXiv '24} &
62M &&
\underline{4.4} & 
\underline{97.9} & &
7.5 & 
94.8 & &
13.2 &
86.2 & &
--- &
--- & &
\textbf{6.5} &
\textbf{95.4}
\\

Metric3D \cite{yin2023metric3dv1}~\ven{ICCV '23} &
8M &&
5.0 &
96.6 & &
\underline{5.8} &
\underline{97.0} & &
\underline{6.4} &
\underline{96.5} & &
\textbf{7.4} &
\textbf{94.1} & &
22.4 &
80.5
\\

Metric3D v2 \cite{hu2024metric3dv2}~\ven{TPAMI '24} &
16M &&
\textbf{4.3} &
\textbf{98.1} & &
\textbf{4.4} &
\textbf{98.2} & &
\textbf{4.2} &
\textbf{98.3} & &
---\textsuperscript{\textdagger} &
---\textsuperscript{\textdagger} & &
13.6 &
89.5
\\

\midrule

Marigold \cite{ke2023marigold}~\ven{CVPR '24} &
74K &&
5.5 & 
96.4 & &
9.9 & 
91.6 & &
6.5 & 
96.0 & &
{6.4} & 
{95.1} & &
30.8 & 
{77.3} \\

GeoWizard \cite{fu2024geowizard}~\ven{ECCV '24} &
278K &&
\textcolor{lightgray}{5.2} & 
\textcolor{lightgray}{96.6} & &
\textcolor{lightgray}{9.7} & 
\textcolor{lightgray}{92.1} & &
\textcolor{lightgray}{6.4} & 
\textcolor{lightgray}{96.1} & &
\textcolor{lightgray}{6.1} & 
\textcolor{lightgray}{95.3} & &
\textcolor{lightgray}{29.7} & 
\textcolor{lightgray}{79.2} \\

\hspace{0.5em} reproduced by \cite{martingarcia2024diffusione2eft} & 
278K && %
{5.7} & 
{96.2} & &
{14.4} &
{82.0} & &
{7.5} & 
{94.3} & &
\underline{6.1} & 
95.8 & &
{31.4} & 
{77.1} \\

DepthFM \cite{gui2024depthfm}~\ven{AAAI '25} &
74K &&

6.0 &       
95.5 & &   
9.1 &     
90.2 & &   
6.5 & 
95.4 & &
6.6 &    
94.9 & & 
22.4 &   
\underline{78.5} \\

E2E FT \cite{martingarcia2024diffusione2eft}~\ven{WACV '25} &
74K &&
\textbf{5.2} & %
\underline{96.6} & &
9.6 & %
\underline{91.9} & &
6.2 &  %
95.9 & &
\textbf{{5.8}} & %
\textbf{96.2} & &
30.2 &  %
77.9 \\

Lotus-G \cite{he2024lotus}~\ven{arXiv '24} &
59K &&
\underline{5.4} & \underline{96.6} &&
11.3 & 87.7 &&
6.2 & 96.1 &&
\underline{6.0} & \underline{96.0} &&
--- & --- \\

\arrayrulecolor{gray!50!white}
\midrule
\arrayrulecolor{black}

\textbf{Diff2Flow} &
74K &&
5.7 & %
\textbf{96.7} & &
\textbf{{8.7}} & %
\textbf{92.0} & &
\textbf{5.5} &  %
\textbf{97.4} & &
6.2 & %
95.7 & &
\textbf{21.6} &  %
\textbf{79.5} \\

\textbf{Diff2Flow} (LoRA) &
74K &&
5.9 & %
96.4 & &
\underline{9.5} & %
90.8 & &
\underline{6.0} &  %
\underline{96.8} & &
6.4 & %
95.5 & &
\underline{22.6} &  %
\underline{78.5} \\

\bottomrule
\end{tabularx}

%% file: tbl/reflow-sota.tex
\footnotesize
\centering
\setlength{\tabcolsep}{2.0pt}
\begin{tabular}{l @{\hskip .1in} c @{\hskip .08in} c @{\hskip .08in} c @{\hskip .08in} c}
\toprule
Method & \#Params $\downarrow$ & Steps & FID $\downarrow$ & CLIP $\uparrow$ \\
\midrule

SDv1.5+DPM Solver \cite{lu2022dpm}~\ven{NeurIPS '22}
    & 0.9B   & 25    & 20.10   & 0.318     \\

\arrayrulecolor{gray!50!white}
\midrule
\arrayrulecolor{black}

Rectified Flow \cite{liu2022rectifiedflow}~\ven{ICLR '23}
    & 0.9B  & 25    & 21.65   & \textbf{0.315}     \\

\textbf{Diff2Flow} (LoRA)
    & 62M    & 25    & \textbf{21.45}   & 0.314   \\

\arrayrulecolor{gray!50!white}
\midrule
\arrayrulecolor{black}

PeRFlow \cite{yan2024perflowpiecewiserectifiedflow}~\ven{arXiv '24}
    &  0.9B  &  4    & \textbf{22.97}   & 0.294     \\

\textbf{Diff2Flow} (LoRA)
    & 62M      & 4    & 25.29   & \textbf{0.313}   \\

\arrayrulecolor{gray!50!white}
\midrule
\arrayrulecolor{black}

Rectified Flow \cite{liu2022rectifiedflow}~\ven{ICLR '23}
    & 0.9B   & 2    & \textbf{31.35}   & 0.296     \\

\textbf{Diff2Flow} (LoRA)
    & 62M      & 2    & 32.31   & \textbf{0.305}   \\

\bottomrule
\end{tabular}

%% file: figures/txt2img/t2i_nfe-cfg.tex
\setlength\tabcolsep{0.1pt}
\center
\small
\newcommand{\imgwidth}{0.12\textwidth}

\newcommand{\imagepng}[1]{
\includegraphics[width=\imgwidth]{figures/txt2img/#1.jpg}
}

\begin{tabular}{ccc}
    \imagepng{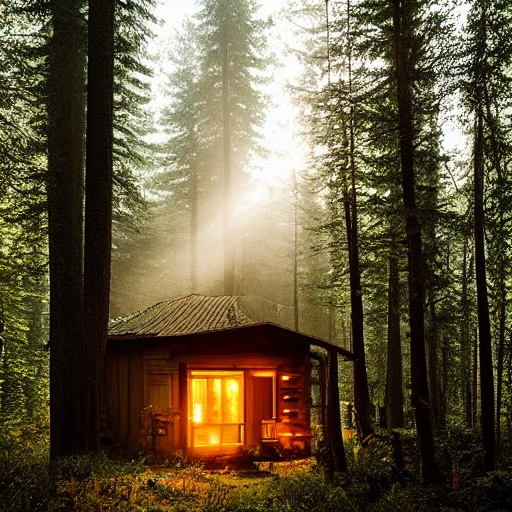} &
    \imagepng{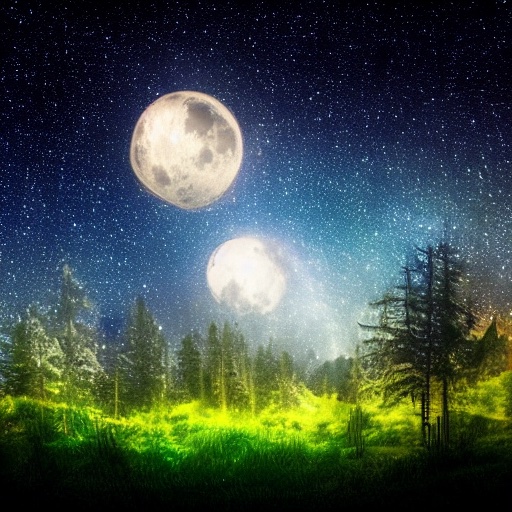} &
    \imagepng{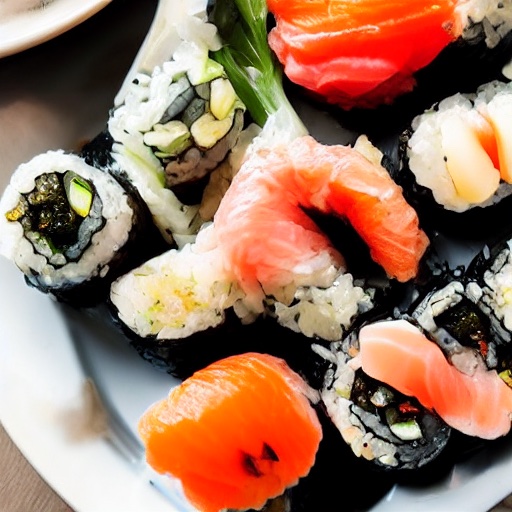} \\
\end{tabular}

%% file: sec/4.3_img2depth.tex
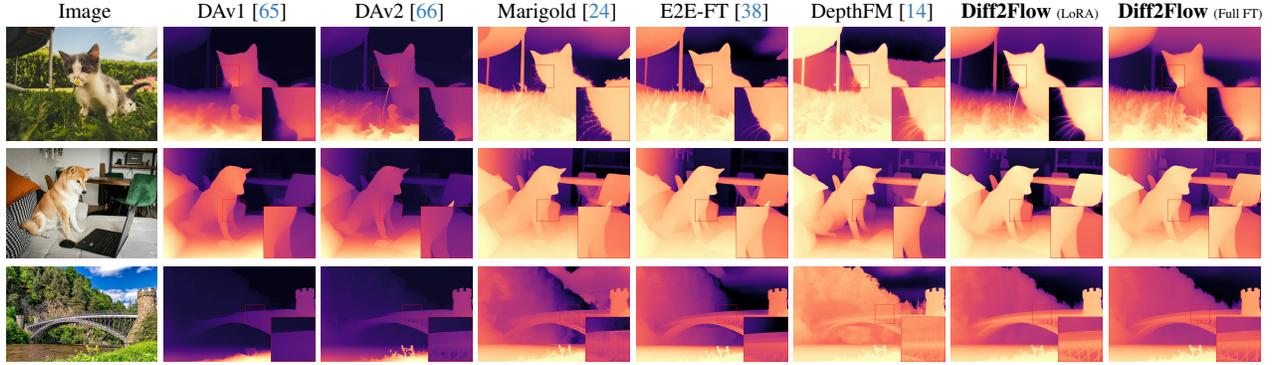
\begin{figure*}[t]
    \input{figures/img2depth/I2D_qualitative}
    \caption{
    Zero-shot qualitative results on real-world imagery. Our methods produce depth predictions with perceptually higher fidelity and finer detail. All predictions are made using the optimal settings of the models. Best viewed when zoomed in.
    }
    \label{fig:i2d:qualitative}
\end{figure*}

\begin{figure*}
    \centering
    \begin{subfigure}[b]{0.45\textwidth}
        \centering
        \includegraphics[height=0.12\textheight, keepaspectratio]{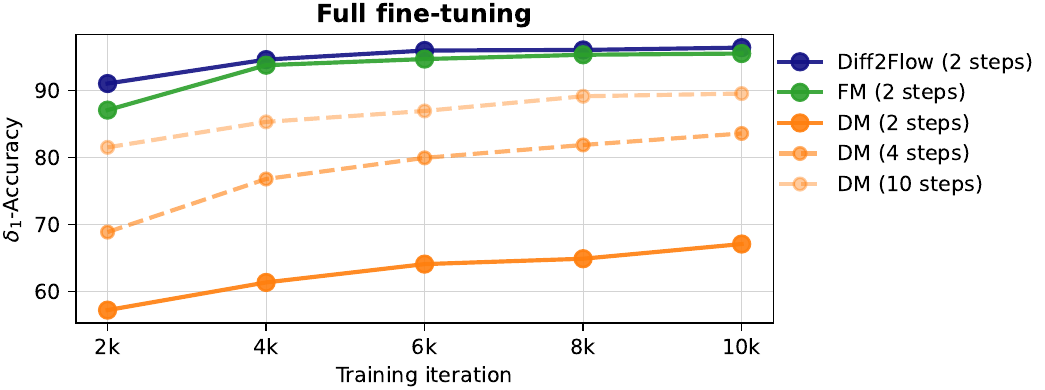}
    \caption{$\qquad \qquad$}
    \label{fig:i2d:dm-fm-obj:full-ft}
    \end{subfigure}
    \qquad
    \begin{subfigure}[b]{0.45\textwidth}
        \centering
        \includegraphics[height=0.12\textheight, keepaspectratio]{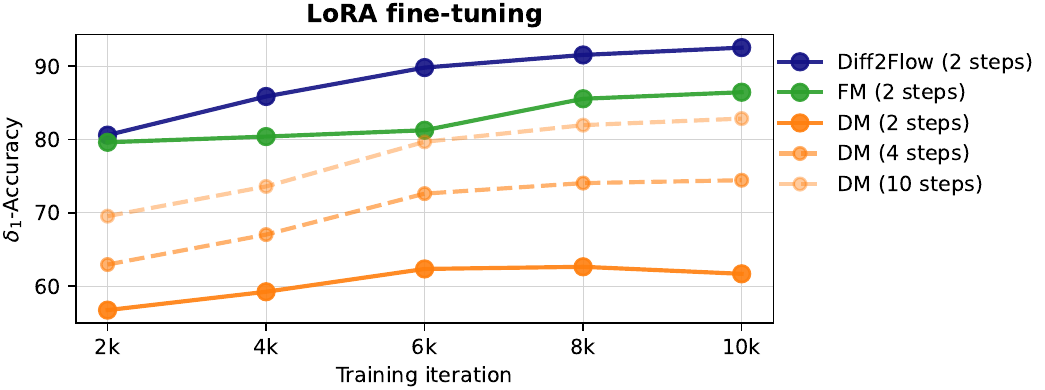}
    \caption{$\qquad \qquad$}
    \label{fig:i2d:dm-fm-obj:lora}
    \end{subfigure}
    \caption{Results on NYUv2 depth benchmark \cite{silberman12nyuv2} with $4$ ensemble members. \textbf{a)} With full fine-tuning we find that FM adapts to the I2D task very quickly, but adding the objective change leads to even quicker convergence and better results. In contrast, the diffusion-based counterpart \cite{ke2023marigold} suffers from slower convergence and requires more inference steps. \textbf{b)} The difference between FM and Diff2Flow is even more pronounced if we limit the capacity of the model by using Low-Rank adaptations. With Diff2Flow, the model does not need to learn to change the objective and can focus only on the task itself.}
    \label{fig:i2d:dm-fm-obj}
\end{figure*}

In addition to common tasks, we also show the versatility of our approach for domain adaptation finetuning. Previous methods have shown strong performance in using large-scale image generative models for affine-invariant monocular depth estimation \cite{ke2023marigold, fu2024geowizard, gui2024depthfm}. \textit{Marigold} \cite{ke2023marigold} directly finetune Stable Diffusion on synthetic image-depth data using the standard $v$-parameterized Diffusion loss (\Cref{eq:v-param}). Following a similar setup, \textit{DepthFM} \cite{gui2024depthfm} propose to convert Stable Diffusion into a flow matching model by training with the flow matching (see \Cref{eq:fm_loss}) loss. This approach requires aligning the model’s outputs to a different objective, interpolant, and timestep scaling (with $t_\text{DM} \in [0, 1000]$ in diffusion versus $t_\text{FM} \in [0, 1]$ in flow matching). This process incurs significant computational cost, limiting the training efficiency and final performance. However, the flow matching nature of DepthFM allows for faster inference, at slightly reduced performance when given enough training. Our method mitigates this problem, accelerating training, and improving performance. The training and evaluation metrics are provided in the Appendix.

\paragraph{Results}
\Cref{tab:i2d:sota} compares our method with both discriminative and generative depth predictors. Trained exclusively on synthetic data, our method matches or outperforms the prior state-of-the-art in monocular depth estimation, as shown in \Cref{fig:i2d:dm-fm-obj}. The diffusion-based Marigold model \cite{ke2023marigold} struggles to achieve high accuracy with a limited number of sampling steps. 
In contrast to both DepthFM and Marigold, our method achieves competitive performance after very few training iterations and with only two sampling steps. The performance difference becomes even more pronounced when training with LoRA, where model capacity is limited. \Cref{fig:i2d:dm-fm-obj:lora} shows that this limitation prevents a straightforward application of the FM loss to the diffusion model. DepthFM struggles here as the model has to learn the adaptation from the $v$-parameterization to the FM objective. The successful adaptation of the diffusion-based image prior to flow matching supports our initial hypothesis that the integration of our objective change facilitates this transformation.
In \Cref{tab:i2d:lora-ablation}, we present an ablation study on performance with different numbers of trainable parameters. Fine-tuning just $1/4$ of the original parameters is sufficient to achieve competitive performance with previous methods, particularly with the fully finetuned DepthFM model.
\Cref{fig:i2d:qualitative} qualitatively compares our method with state-of-the-art depth estimators on in-the-wild images. While discriminative methods often lack detail and fidelity, generative models capture fine details and avoid the averaging effect common to discriminative approaches.

\begin{table}[t]
    \input{tbl/i2d-lora-ablation}
    \caption{Zero-shot evaluation on NYUv2 \cite{silberman12nyuv2} for different number of trainable parameters. We can achieve competitive performance with only a fraction of previous SOTA methods.}
  \label{tab:i2d:lora-ablation}
\end{table}

%% file: figures/img2depth/I2D_qualitative.tex
\setlength\tabcolsep{0.1pt}
\center
\small
\newcommand{\imgwidth}{0.115\textwidth}

\newcommand{\imagepng}[2]{
\includegraphics[width=\imgwidth]{figures/img2depth/qualitative_jpg/#1_#2.jpg}
}

\newcommand{\rowdepth}[1]{
    \imagepng{#1}{image} &
    \imagepng{#1}{DepthAnythingv1_zoomed_in} &
    \imagepng{#1}{DepthAnythingv2_zoomed_in} &
    \imagepng{#1}{Marigold_zoomed_in} &
    \imagepng{#1}{E2E-FT_zoomed_in} &
    \imagepng{#1}{DepthFM_zoomed_in} &
    \imagepng{#1}{Ours_LoRA_zoomed_in} &
    \imagepng{#1}{Ours_zoomed_in}
    
}

\begin{tabular}{cccccccc}
    \footnotesize Image &
    \footnotesize DAv1 \cite{depth_anything_v1} &
    \footnotesize DAv2 \cite{depth_anything_v2} &
    \footnotesize Marigold \cite{ke2023marigold} &
    \footnotesize E2E-FT \cite{martingarcia2024diffusione2eft} &
    \footnotesize DepthFM \cite{gui2024depthfm} &
    \footnotesize \textbf{Diff2Flow} \tiny (LoRA) &
    \footnotesize \textbf{Diff2Flow} \tiny (Full FT)
    \\
    \rowdepth{00} \\
    \rowdepth{01} \\
    \rowdepth{06} \\
\end{tabular}

%% file: tbl/i2d-lora-ablation.tex
\footnotesize
\centering
\setlength{\tabcolsep}{2.0pt}
\begin{tabular}{@{}lc@{\hskip .3in}c@{\hskip .3in}c@{}}
\toprule
Method & \#Params $\downarrow$ & AbsRel $\downarrow$ & $\delta_1$ $\uparrow$ \\
\midrule
Marigold + E2E FT \cite{martingarcia2024diffusione2eft}
    & $866$M & 5.2 & 96.6 \\
\arrayrulecolor{gray!50!white}
\midrule
\arrayrulecolor{black}
Diff2Flow      & $866$M  & 5.7  & 96.7 \\
Diff2Flow \tiny (LoRA base)  & $222$M  & 5.9  & 96.4 \\
Diff2Flow \tiny (LoRA small) & $62$M   & 6.9  & 95.0 \\
\bottomrule
\end{tabular}

%% file: sec/5_conclusion.tex
We introduce a novel framework, \textit{Diff2Flow}, that effectively converts pre-trained diffusion into flow matching models, providing a twofold optimization.
First, this approach allows us to efficiently use the diffusion model's prior by exploiting its powerful generative capabilities without retraining from scratch.
Second, by adapting it to the flow matching paradigm, we gain key benefits unique to flow matching: fast inference suitable for downstream tasks, and the ability to straighten sampling trajectories through techniques such as reflow, which further improves efficiency and performance.
As a result, \textit{Diff2Flow} not only improves training and inference efficiency but also delivers competitive performance in diverse tasks, including text-to-image synthesis and monocular depth estimation. By using parameter-efficient fine-tuning methods such as LoRA, our framework further minimizes computational requirements and demonstrates a practical, scalable way to merge diffusion and flow matching paradigms in generative modeling.

%% file: sec/X_suppl.tex
\setcounter{section}{0}
\onecolumn
\clearpage
\appendix

\begin{center}
    \part*{Diff2Flow: Training Flow Matching Models via Diffusion Model Alignment}
\end{center}

\begin{figure*}[h]
\centering
\includegraphics[width=\linewidth]{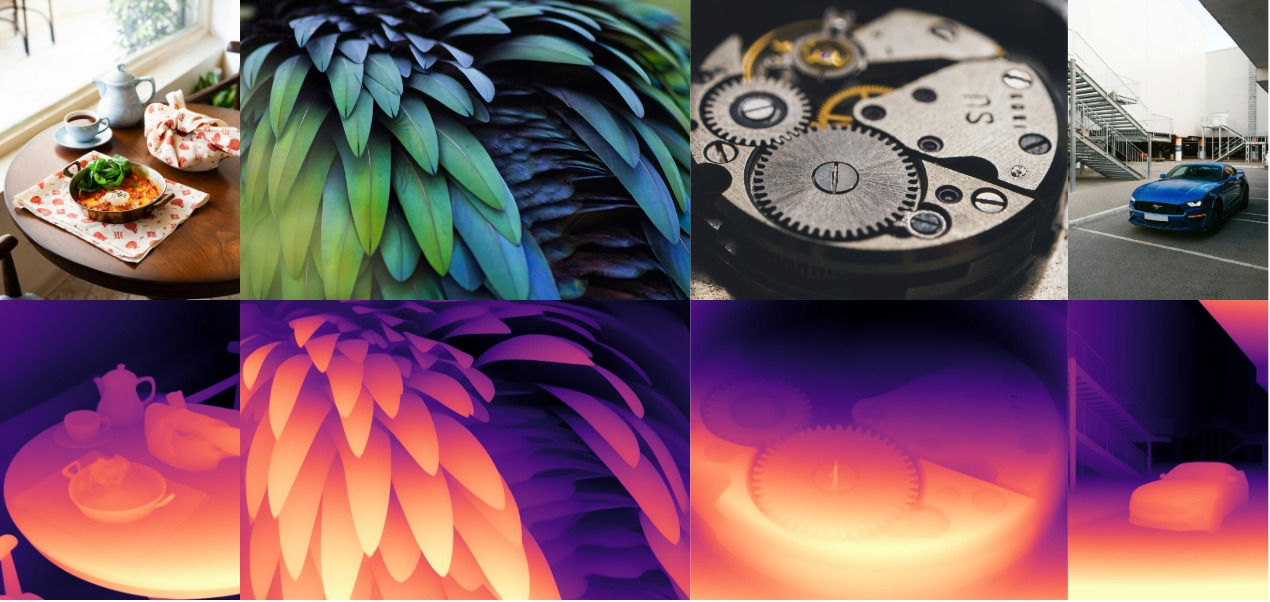}
\caption{Diff2Flow enables fast monocular depth estimation with high fidelity.}
\end{figure*}

\section{Implementation Details}
\label{sup:sec:implementation}

\paragraph{Text-to-Image.} For the text-to-image task, we fine-tune Stable Diffusion 2.1, aligning its $\mathbf{v}$-parameterization to Flow Matching. For the comparison, we fine-tune three models: the diffusion baseline, the diffusion model with Flow Matching loss, and our proposed Diff2Flow adaptation. Each model is trained for $20$k iterations using a constant learning rate of  $1 \times 10^{-5}$ and a batch size of $64$ on the LAION-Aesthetics dataset \cite{schuhmann2022laion}, which contains high-aesthetic-score images paired with synthetically generated captions. We evaluate all models on the COCO 2017 dataset \cite{lin2014microsoft_coco2017} using ODE sampling. In our Low-Rank Adaptation (LoRA) setup, we set the rank of convolutions and attention layers to $20$\% of each layer’s respective feature dimension. This configuration results in a model with 222M trainable parameters for the LoRA version, compared to the 866M parameters of the original Stable Diffusion 2.1 model.

\paragraph{Reflow.} Adapting Stable Diffusion with our proposed method, allows us to perform rectification of sampling trajectories, as proposed in \cite{liu2022rectifiedflow} for Flow Matching models. Rectification relies on pre-computed image-noise pairs, which we generate by sampling approximately 1.8M images with a classifier-free guidance scale of $7.5$, using prompts from the LAION-Aesthetics dataset \cite{schuhmann2022laion} and $40$ sampling steps. We then perform 1-rectification training for 60k gradient updates on these image-noise pairs. We fix the LoRA rank to $64$ across all convolutional, self-attention, and feedforward layers, resulting in a model with a total of 62M trainable parameters. We train this model with a batch size of $128$ and a decaying learning rate schedule starting from $2 \times 10^{-5}$, and evaluate it on the COCO 2017 dataset \cite{lin2014microsoft_coco2017}.

\paragraph{Image-to-Depth.} 
We follow the training paradigm of \cite{ke2023marigold, gui2024depthfm} and use a mixture of Hypersim~\cite{roberts2021hypersim} and Virtual Kitti v2~\cite{cabon2020vkitti2} data. Similar to~\cite{gui2024depthfm} we log-normalize the depth data, as we found it to make better use of the input data space.
We evaluate zero-shot on five benchmark datasets: NYUv2~\cite{silberman12nyuv2}, DIODE~\cite{diode_dataset}, ScanNet~\cite{dai2017scannet}, KITTI~\cite{geiger2013kitti}, and ETH3D~\cite{schops2017multiEth3d}. We use the evaluation suite from \cite{ke2023marigold} and align an ensemble of estimated depth maps to the ground truth depth with least squares fitting. We report the average relative difference between the ground-truth depth and the aligned predicted depth at each pixel (AbsRel), as well as $\delta_1$-Accuracy, which is the percentage of pixels for which the ratio between the aligned predicted depth and the ground-truth depth is below $1.25$. 
Similar to Marigold \cite{ke2023marigold}, we train for 20k gradient updates with a batch size of 32 and a decaying learning rate schedule. For LoRA fine-tuning, we explore two variants: the first, “LoRA base” sets the rank to $20$\% of the respective feature dimension for all convolutional and attention layers, resulting in 222M trainable parameters. The second, a smaller LoRA model, fixes the rank to 64 across all convolutional, self-attention, and feedforward layers, resulting in 62M trainable parameters. We train our models on a resolution of $384 \times 512$. During evaluation, we resize the images to this size and subsequently resize our depth prediction to the ground truth resolution. We evaluate our models using an ensemble size of four and 10 sampling steps.

\section{Qualitative Results}
\label{sup:sec:qualitative}

\subsection{Reflow}

\begin{figure*}
    \centering
    \includegraphics[width=0.9\linewidth]{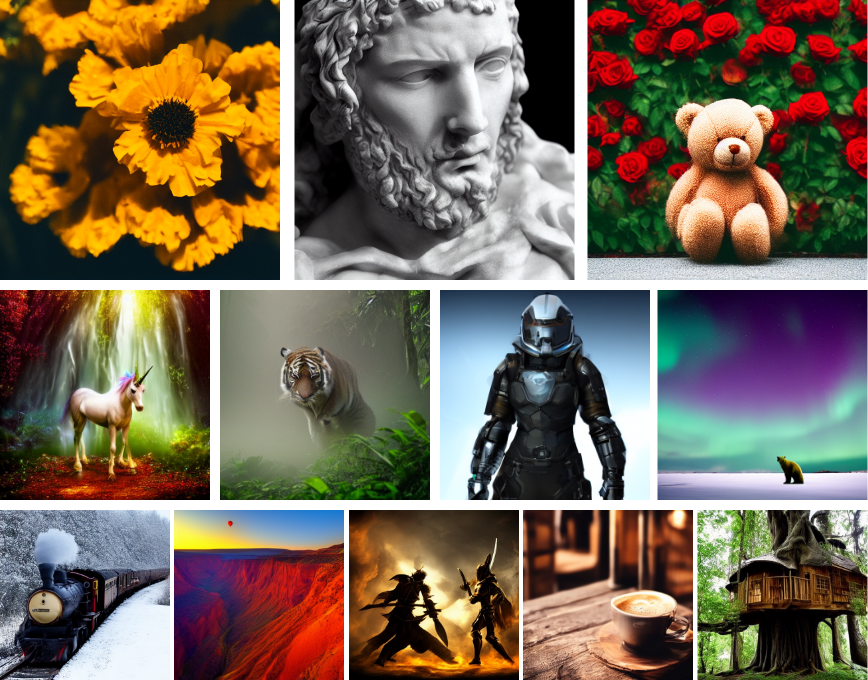}
    \caption{4-step inference results of our Diff2Flow-reflow model, using Stable Diffusion 1.5 as the prior diffusion model}
    \label{supp:fig:4stepreflow}
\end{figure*}

\begin{figure}
    \centering
    \includegraphics[width=0.9\linewidth]{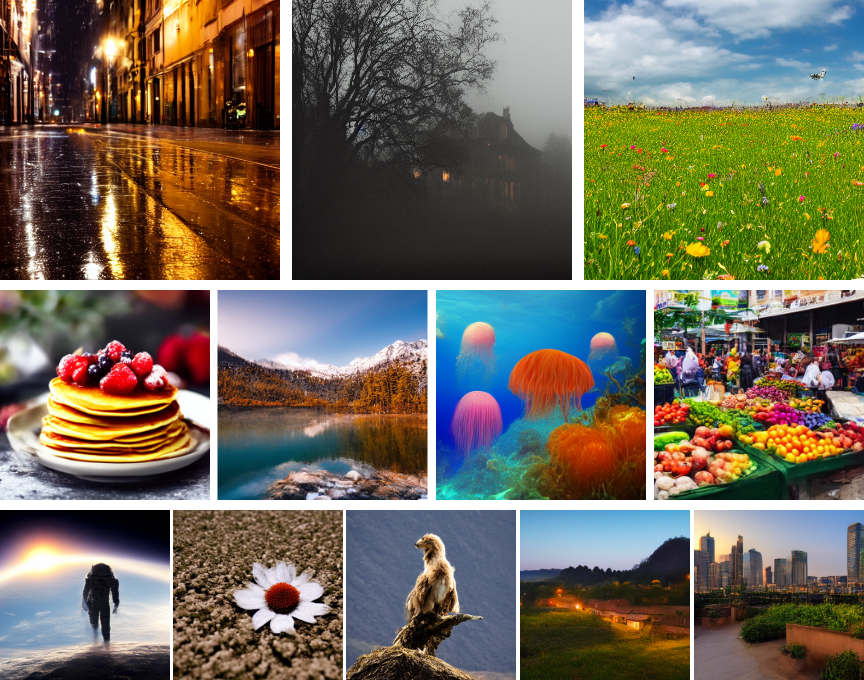}
    \caption{2-step inference results of our Diff2Flow-reflow model, using Stable Diffusion 1.5 as the prior diffusion model}
    \label{supp:fig:2stepreflow}
\end{figure}

In addition to the samples presented in \cref{fig:reflow:cfg-ablation}, we provide further qualitative results in \cref{supp:fig:4stepreflow} and \cref{supp:fig:2stepreflow}. By applying only the first rectified flow, Diff2Flow significantly reduces the number of diffusion generation steps while maintaining competitive performance compared to state-of-the-art flow matching approaches.

\subsection{Image-to-Depth}

\begin{figure*}[t]
    \centering
    \includegraphics[width=.95\linewidth]{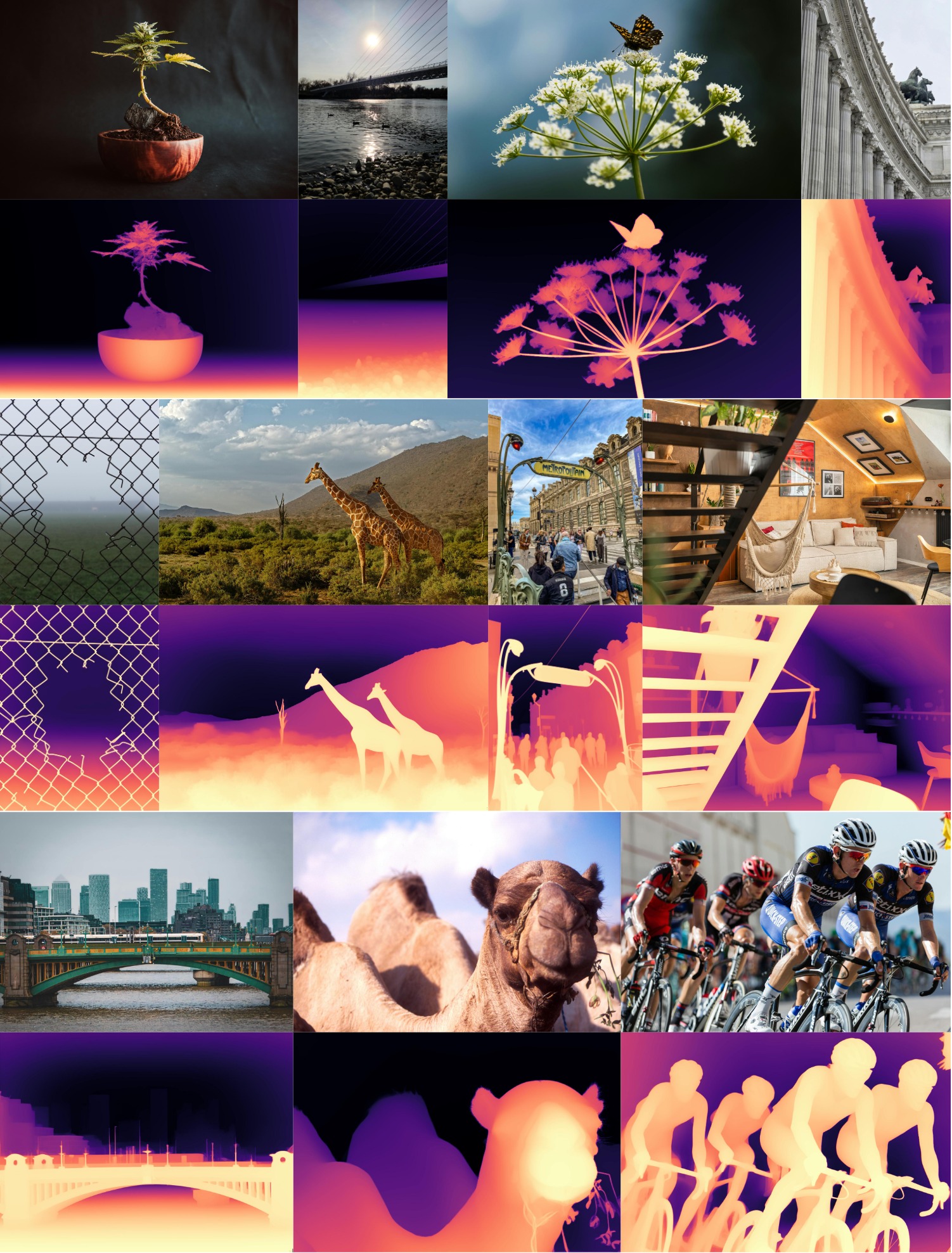}
    \caption{Qualitative results for monocular depth prediction.}
    \label{fig:i2d:qualitative:supp3}
\end{figure*}

In addition to the examples shown in \cref{fig:i2d:qualitative}, we present additional qualitative comparisons for our monocular depth estimation in \cref{fig:i2d:qualitative:page1}. Our method consistently generates depth predictions with perceptually higher fidelity and finer details compared to the state-of-the-art models. \cref{fig:i2d:qualitative:supp3} shows additional depth estimations of our Diff2Flow depth estimation model for in-the-wild images.

\begin{figure*}[t]
    \centering
    \begin{subfigure}{\textwidth}
        \centering
        \input{figures/img2depth/I2D_supp}
        \label{fig:i2d:qualitative:supp}
    \end{subfigure}
    \caption{More qualitative results for monocular depth prediction compared to the state-of-the-art models (Part 1).}
    \label{fig:i2d:qualitative:page1}
\end{figure*}
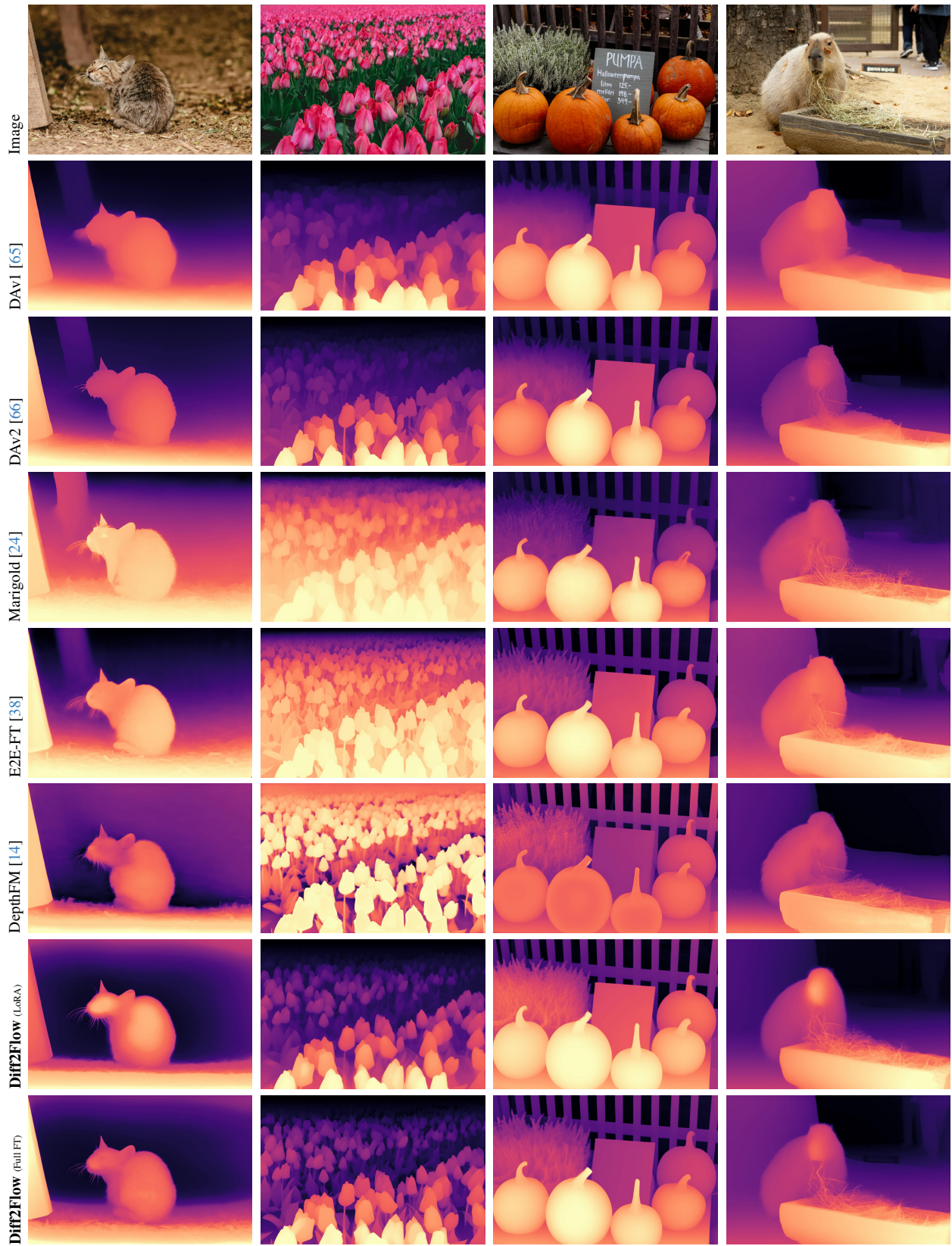

\begin{figure*}[t]
    \ContinuedFloat %
    \centering
    \begin{subfigure}{\textwidth}
        \centering
        \input{figures/img2depth/I2D_supp2}
        \label{fig:i2d:qualitative:supp2}
    \end{subfigure}
    \caption{More qualitative results for monocular depth prediction compared to the state-of-the-art models (Part 2).}
    \label{fig:i2d:qualitative:page2}
\end{figure*}

%% file: figures/img2depth/I2D_supp.tex
\setlength\tabcolsep{0.1pt}
\centering
\small
\newcommand{\imgheight}{0.15\textwidth}

\newcommand{\imagepng}[2]{
\includegraphics[height=\imgheight]{figures/img2depth/supp_images_jpg_resized/#1_#2.jpg}
}
\newcommand{\imagejpg}[2]{
\includegraphics[height=\imgheight]{figures/img2depth/supp_images_jpg_resized/#1_#2.jpg}
}

\newcommand{\wrapperpng}[2]{
    \foreach \n in {#2} {\imagepng{#1}{\n}}
}

\newcommand{\wrapperjpg}[2]{
    \foreach \n in {#2} {\imagejpg{#1}{\n}}
}

\begin{tabular}{c c c c c}
    \rotatebox{90}{\footnotesize Image} &
    \wrapperjpg{image}{0,1,pumpkins,capybara} \\
    
    \rotatebox{90}{\footnotesize DAv1 \cite{depth_anything_v1}} &
    \wrapperpng{DepthAnythingv1}{0,1,pumpkins,capybara} \\
    \rotatebox{90}{\footnotesize DAv2 \cite{depth_anything_v2}} &
    \wrapperpng{DepthAnythingv2}{0,1,pumpkins,capybara} \\
    \rotatebox{90}{\footnotesize Marigold \cite{ke2023marigold}} &
    \wrapperpng{Marigold}{0,1,pumpkins,capybara} \\
    \rotatebox{90}{\footnotesize E2E-FT \cite{martingarcia2024diffusione2eft}} &
    \wrapperpng{E2E-FT}{0,1,pumpkins,capybara} \\
    \rotatebox{90}{\footnotesize DepthFM \cite{gui2024depthfm}} &
    \wrapperpng{DepthFM}{0,1,pumpkins,capybara} \\
    \rotatebox{90}{\footnotesize \textbf{Diff2Flow} \tiny (LoRA)} &
    \wrapperpng{Ours_LoRA}{0,1,pumpkins,capybara} \\
    \rotatebox{90}{\footnotesize \textbf{Diff2Flow} \tiny (Full FT)} &
    \wrapperpng{Ours}{0,1,pumpkins,capybara} \\
\end{tabular}

%% file: figures/img2depth/I2D_supp2.tex
\setlength\tabcolsep{0.1pt}
\centering
\small
\newcommand{\imgheight}{0.15\textwidth}

\newcommand{\imagepng}[2]{
\includegraphics[height=\imgheight]{figures/img2depth/supp_images_jpg_resized/#1_#2.jpg}
}
\newcommand{\imagejpg}[2]{
\includegraphics[height=\imgheight]{figures/img2depth/supp_images_jpg_resized/#1_#2.jpg}
}

\newcommand{\wrapperpng}[2]{
    \foreach \n in {#2} {\imagepng{#1}{\n}}
}

\newcommand{\wrapperjpg}[2]{
    \foreach \n in {#2} {\imagejpg{#1}{\n}}
}

\begin{tabular}{c c c c c}
    \rotatebox{90}{\footnotesize Image} &
    \wrapperjpg{image}{4,horse,7,biscuit,wires} \\
    
    \rotatebox{90}{\footnotesize DAv1 \cite{depth_anything_v1}} &
    \wrapperpng{DepthAnythingv1}{4,horse,7,biscuit,wires} \\
    \rotatebox{90}{\footnotesize DAv2 \cite{depth_anything_v2}} &
    \wrapperpng{DepthAnythingv2}{4,horse,7,biscuit,wires} \\
    \rotatebox{90}{\footnotesize Marigold \cite{ke2023marigold}} &
    \wrapperpng{Marigold}{4,horse,7,biscuit,wires} \\
    \rotatebox{90}{\footnotesize E2E-FT \cite{martingarcia2024diffusione2eft}} &
    \wrapperpng{E2E-FT}{4,horse,7,biscuit,wires} \\
    \rotatebox{90}{\footnotesize DepthFM \cite{gui2024depthfm}} &
    \wrapperpng{DepthFM}{4,horse,7,biscuit,wires} \\
    \rotatebox{90}{\footnotesize \textbf{Diff2Flow} \tiny (LoRA)} &
    \wrapperpng{Ours_LoRA}{4,horse,7,biscuit,wires} \\
    \rotatebox{90}{\footnotesize \textbf{Diff2Flow} \tiny (Full FT)} &
    \wrapperpng{Ours}{4,horse,7,biscuit,wires} \\
\end{tabular}